\documentclass[runningheads]{llncs}
\usepackage[colorlinks]{hyperref}
\usepackage{graphicx}
\usepackage{makecell}
\usepackage{amsmath}
\usepackage{booktabs}
\usepackage{array}
\usepackage{tabularx}
\usepackage{multirow}
\usepackage{amssymb}
\usepackage{algorithm}
\usepackage{algorithmic}
\usepackage{tabularray}
\usepackage{graphicx}
\usepackage{setspace}

\usepackage{color}
\usepackage{soul}
\usepackage{tikz}
\usepackage{marvosym}
\usetikzlibrary{matrix}
\definecolor{tomato}{rgb}{1.0, 0.39, 0.28}
\usepackage{xargs} 

\begin{document}
\title{SDF-Net: A Hybrid Detection Network for Mediastinal Lymph Node Detection on Contrast CT Images}
\titlerunning{SDF-Net for Lymph Node Detection}

\author{Jiuli Xiong \inst{1, 2}
\and Lanzhuju Mei \inst{1}
\and Jiameng Liu \inst{1}
\and Dinggang Shen \inst{1, 2, 3} \textsuperscript{(\Letter)}
\and Zhong Xue \inst{2}
\and Xiaohuan Cao \inst{2} \textsuperscript{(\Letter)}
}
\authorrunning{Jiuli Xiong et al.}

\institute{School of Biomedical Engineering $\&$ State Key Laboratory of Advanced Medical Materials and Devices, ShanghaiTech University, Shanghai 201210, China  \\
\email{dinggang.shen@gmail.com} 
\and Department of Research and Development, United Imaging Intelligence, \\Shanghai 200230, China \\
\email{xiaohuan.cao@uii-ai.com}
\and Shanghai Clinical Research and Trial Center, Shanghai 201210, China} 
\maketitle  
\begin{abstract}

Accurate lymph node detection and quantification are crucial for cancer diagnosis and staging on contrast-enhanced CT images, as they impact treatment planning and prognosis. However, detecting lymph nodes in the mediastinal area poses challenges due to their low contrast, irregular shapes and dispersed distribution. In this paper, we propose a Swin-Det Fusion Network (SDF-Net) to effectively detect lymph nodes. SDF-Net integrates features from both segmentation and detection to enhance the detection capability of lymph nodes with various shapes and sizes. Specifically, an auto-fusion module is designed to merge the feature maps of segmentation and detection networks at different levels. To facilitate effective learning without mask annotations, we introduce a shape-adaptive Gaussian kernel to represent lymph node in the training stage and provide more anatomical information for effective learning. Comparative results demonstrate promising performance in addressing the complex lymph node detection problem.

\end{abstract}

\section{Introduction}

\begin{figure}[h]
  \centering
  \includegraphics[width=0.65\textwidth]{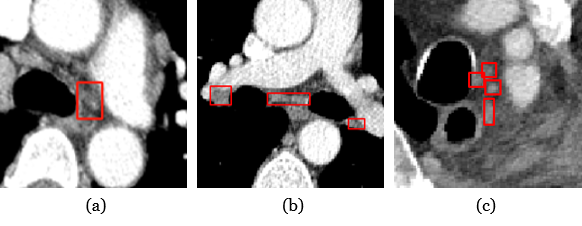}
  \caption{Typical lymph nodes with annotated bounding boxes in contrast-enhanced CT images. (a) The lymph node exhibits low contrast with the surrounding tissues; (b) the lymph nodes have diverse size and shape; (c) the lymph nodes often clustered together.}
  \label{nodesvision}
\end{figure}

Accurate assessment of lymph node (LN) metastasis is crucial for the diagnosis and staging of cancer \cite{bouget2023mediastinal}, and impact treatment planning and prognosis. Therefore, precise and automatic detection and quantification of lymph nodes from contrast-enhanced CT images is clinically imperative. However, accurately detecting lymph nodes on CT images poses several challenges. First, lymph nodes have lower contrast compared to surrounding tissues, making them difficult to be separated. While contrast CT can help differentiate vessels, identifying surrounding soft tissues remains difficult. Second, lymph nodes often exhibit irregular shapes and variable sizes, particularly in case of metastasis. They are also dispersed throughout the mediastinal area, rather than concentrated in a specific organ, making detection more challenging. Third, some abnormal lymph nodes may cluster together, further complicating accurate detection. Fig. \ref{nodesvision} illustrates some typical examples of mediastinal lymph nodes.

Deep learning-based object detection algorithms \cite{ren_faster-rcnn_2016,liu_ssd_2016,redmon_yolov3_2018,law_cornernet_nodate,zhou2019objects}, can be categorized as \textit{anchor-based} \cite{ren_faster-rcnn_2016,redmon_yolov3_2018,liu_ssd_2016} and \textit{anchor-free} \cite{law_cornernet_nodate,tian_fcos_2019,zhou2019objects,redmon_yolov3_2018} methods. Popular \textit{anchor-based} algorithms include Faster R-CNN \cite{ren_faster-rcnn_2016}, SSD \cite{liu_ssd_2016} and YOLO \cite{redmon_yolov3_2018}. These methods require a series of predefined anchors, and the deep network is applied to learn the relations between the targets and these anchors. They have shown promising results when the shape and size of detection target is close to the predefined anchors. But it may be hard to tackle the lymph nodes detection problem due to diverse shapes and sizes. On the other hand, \textit{anchor-free} detection methods like CenterNet \cite{zhou2019objects}/CornerNet 
 \cite{law_cornernet_nodate} and transformer-based algorithms (e.g., DETR \cite{carion_end--end_2020-detr} and its variations \cite{zhu_deformable_2021}) are currently more popular. In anchor-free frameworks, targets are typically represented by a center point and a bounding box (bbox), allowing the network to directly learn the bbox coordinates without predefined anchors. However, these methods may not perform well in small target detection, and for medical images, small target or lesion detection is more meaningful clinically. 

Besides object detection techniques, segmentation methods have shown the efficacy for lesion detection and quantification \cite{isensee_nnu-net_2021,hatamizadeh_swin_2022-swin-unetr,ariji_segmentation_lymph-2022}. Lesions or the targets are represented by the delineation masks, and the learning process can be more effective compared with using a center point or bounding box to represent the lesions. It can obviously enhance the small lesion detection, but the clustered lesions are difficult to separate since the target center and boundary are treated equally. Additionally, in real-world scenarios, obtaining accurate lesion delineation in 3D medical images is difficult due to the huge workload and blurred lesion boundaries.

In this study, we propose a hybrid detection algorithm, called Swin-Det Fusion Network (SDF-Net), for addressing the challenges of lymph node detection. Our approach integrates the complementary advantages of  segmentation methods and adopts a weakly-supervised training process, eliminating the need for labor-intensive annotations. The main contributions can be summarized as follows:

\begin{itemize}
    \item SDF-Net integrates features obtained from segmentation and detection paths, rather than directly fusing final results. The detection path follows an anchor-free approach, incorporating an auto-fusion module to non-linearly merge features of the segmentation path. This design can effectively enhance the detection capability of lymph nodes with various shapes and sizes.
    \item A shape-adaptive Gaussian kernel represents the lymph node in the detection training stage. It is generated automatically from annotated 3D bboxes. This Gaussian kernel can enrich anatomical information of lymph nodes, facilitating more effective learning within the anchor-free detection framework. Additionally, a pseudo-mask is derived from this kernel to enable the training of segmentation path without precise lymph nodes delineation.
\end{itemize}

\section{Methods}

\begin{figure}[h]
  \centering
  \includegraphics[width=1.0\textwidth]{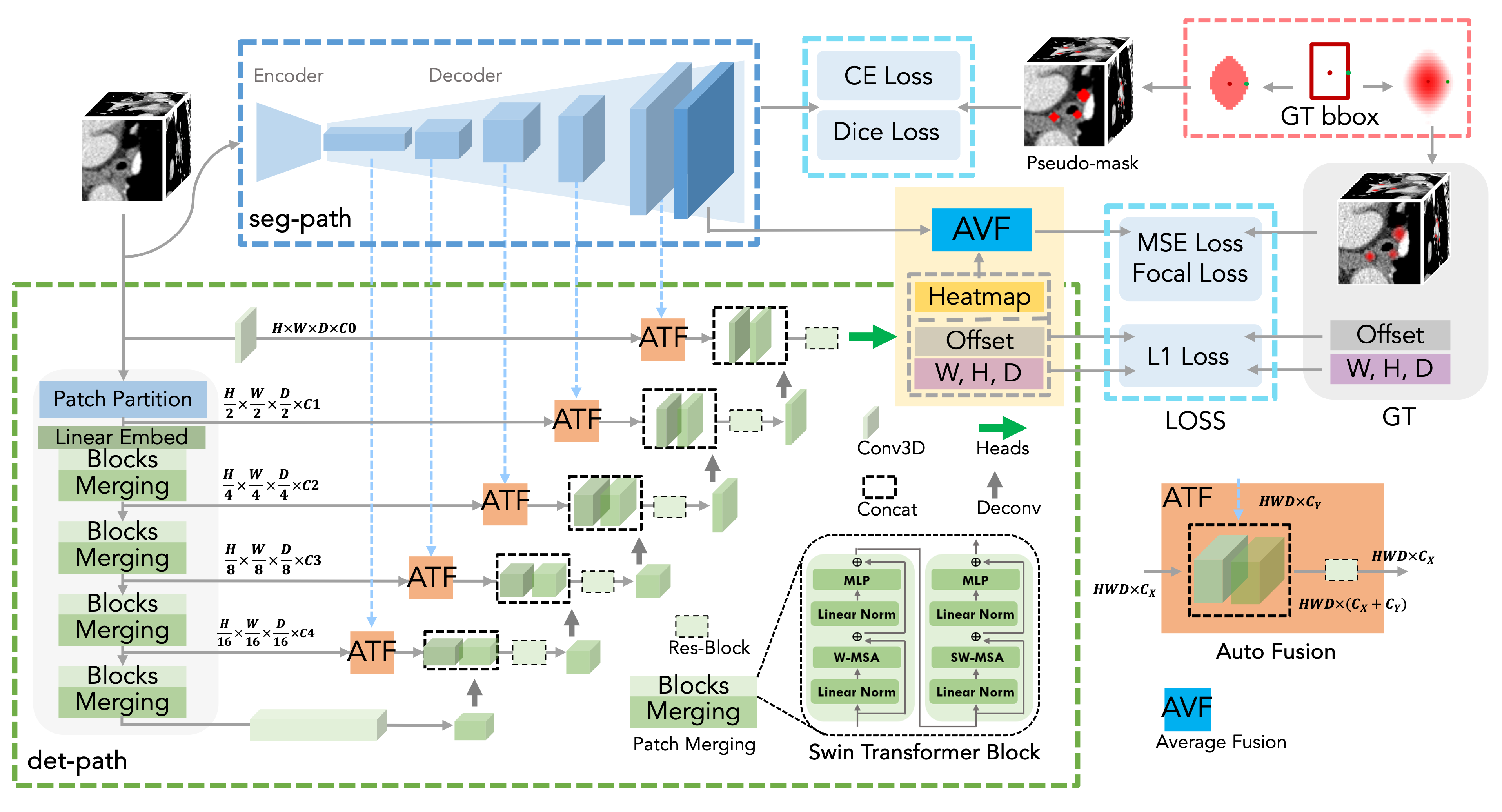}
  \caption{The proposed \textbf{SDF-Net}.}
  \label{modelfig}
\end{figure}

Fig. \ref{modelfig} illustrates the framework of our proposed \textbf{SDF-Net} for LN detection. The original annotation of LN is a 3D bbox. A shape-adaptive 3D Gaussian kernel is generated from bbox to guide the model training. The proposed method involves two pathways for detecting LNs: 1) the \textbf{segmentation path} (seg-path) focuses on learning the pseudo-mask of the target LN, and 2) the anchor-free \textbf{detection path} (det-path) aims to learn the position and bbox information. Initially, the seg-path is trained independently, where the Gaussian kernel is used to generate a pseudo-mask that guides the training of the segmentation model. It also learns high-level feature representations for image and LN interpretation. The det-path is substantially trained by integrating the feature maps from seg-path using the proposed \textbf{auto fusion (ATF)} module. The output of det-path consists 7 channels: 1 channel for Gaussian heatmap; 3 channels for the width, height and depth (W, H, D) of bbox, and 3 channels for the offsets to adjust the target position. During inference, images are fed to the seg-path first and then passed through det-path. The final probability maps from both paths are combined to get detection result. The details of each part will be introduced below.

\subsection{Swin-Det Fusion Network}

To effectively address the issues of diverse shapes of LNs and their clustered appearance, we propose a hybrid network called SDF-Net. This network uses the auxiliary feature information from segmentation model to facilitate the detection performance  of swin-Det model, which is the det-path. The seg-path provides strong guidance for distinguishing LNs from surrounding soft tissues and offers benefits for small LNs identification. The det-path can further enhance the sensitivity and accuracy in LN detection by incorporating high-level features of seg-path, while reducing false positive (FP) detections.

\textbf{Seg-path.} The seg-path is derived from a typical segmentation framework, nnU-Net \cite{isensee_nnu-net_2021}. It is defined in 3D format including 4 levels for encoding and decoding. The detailed architecture can be found in U-Net \cite{ronneberger_u-net_2015}. Since the precise LN mask is difficult to obtain clinically, to train the seg-path, a pseudo-mask is generated by the 3D shape-adaptive Gaussian kernel, where the $\sigma$ of Gaussian is related with W, H, D of the annotated bbox in 3 directions, as shown in the up-right of Fig. \ref{modelfig}. Unlike learning a single center point, pseudo-mask learning can enhance the anatomical and morphological details of LN, which can effectively help to distinguish LN from surrounding tissues and capture more informative features. The loss function of seg-path combines cross-entropy loss $L_{CE} $ and Dice loss $L_{Dice} $ \cite{li2019dice}. Downsampled ground-truth segmentation masks corresponding to hidden layers are used for loss calculation by connecting hidden layers to additional convolution layers to generate coarse outputs, which are then compared to the downsampled masks \cite{wang2020multilay}. The final loss function $L_{seg}$ is defined as:  
\begin{equation}
L_{seg} = \sum_{i=1}^{4} \left (\alpha _{i} \cdot L_{CE}  \left ( P^i_{mask}, G^i_{mask} \right ) + \beta  _{i} \cdot L_{Dice}  \left ( P^i_{mask},G^i_{mask} \right ) \right ),
\end{equation}
where $P^i_{mask}$ is the predicted mask in layer $i$, $G^i_{mask}$ is the pseudo-mask, obtained using Gaussian kernel, of layer $i$, and $\alpha_{i}$ and $\beta_{i}$ are the weights to balance the Dice and cross-entropy losses.

\textbf{Det-path.} The det-path is derived from Swin-UNETR \cite{hatamizadeh_swin_2022-swin-unetr}. Its self-attention mechanism can adopt more complicated detection compared to CenterNet \cite{zhou2019objects}, a typical anchor-free detection framework. To improve the performance of small target detection, two innovative designs are introduced in Swin-UNETR: 1) \textbf{ATF} module to integrate auxiliary features from the seg-path; and 2) the improved design detection output derived from CenterNet \cite{zhou2019objects}. The det-path is trained by fusing high-level features of seg-path using the proposed ATF module, as shown in the orange box of Fig. \ref{modelfig}. The input of ATF is the encoder feature map of det-path and the decoder features of seg-path. They are concatenated and fed into a non-linear learning layer (a residual block), then the output of ATF is concatenated with features of det-path’s decoder for further training. The ATF module can gradually incorporate segmentation features into detection network to enhance detection capability.

The output of det-path has 7 channels to represent the bbox. The last 6 channels include W, H, D and offsets. For the first channel, the heatmap is generated using a shape-adaptive Gaussian kernel rather than a single point and is formulated as a regression learning problem. This can help characterize the transition from the center to the boundary, even the LN boundary is not quite clear. Furthermore, it helps maintain a balance between positive and negative samples during model training. The overall loss function $L_{det}$ of det-path can be defined as:
\begin{equation}
\begin{aligned}
    L_{det} = & \sigma_{1} \cdot L_m(P_{\text{hmap}}, G_{\text{hmap}}) + \sigma_{2} \cdot L_f(P_{\text{hmap}}, G_{\text{hmap}}) \\
           & + \kappa \cdot L_1(P_{\text{Offsets}}, G_{\text{Offsets}}) + \mu  \cdot L_1(P_{\text{WHD}}, G_{\text{WHD}}),
\end{aligned}
\end{equation}
where $P_{\text{hmap}}$ is the predicted heatmap, $G_{\text{hmap}}$ is the Gaussian kernel of ground truth lymph nodes. $P_{\text{Offsets}}$ is the predicted offsets map, $G_{\text{Offsets}}$ is the ground truth computed based on LN center offsets. $P_{\text{WHD}}$ represents the predicted width, height, and depth maps, and $G_{\text{WHD}}$ is the ground truth. $L_{m}$ is MSE loss, $L_{f}$ is focal loss, and $L_{1}$ is L1 loss. $\sigma_{1}$, \(\mathrm{\sigma}_{2}\), $\kappa$ and $\mu$ are the weights for different losses.
Here, both MSE and focal loss \cite{lin_focal_nodate} are used to learning not only local intensity but also shape information.

The advantages of hybrid segmentation and detection in feature level contribute to a comprehensive representation of image content. This synergy can enhance the object localization and semantic understanding, thereby boosting the overall performance of LN detection.

\subsection{Inference Process}

During inference, a CT image is divided into 3D patches with a $15$mm overlapping sliding window. Each patch is processed by the seg-path to generate segmentation probability and feature maps, then passed through the det-path. The probability maps from both paths are merged using the average fusion (AVF) module to produce the final detection heatmap. After obtaining the det-path heatmap, we use the AVF to average the probability map from the segmentation path, leveraging the strengths of both paths. When both paths agree, the result is enhanced; when they disagree, the evaluated twice area has a more robust judgment.
LN centers are determined by locating peaks within a 9$\times$9$\times$9 region that exceeds a predefined threshold, and W, H, and D are then calculated. Non-Maximum Suppression (NMS) \cite{neubeck2006efficient-nms} is applied to remove overlapping detections, yielding the final results.

\section{Results}

\subsection{Dataset and Implementation Details}

The dataset consists of $1070$ subjects, each with a contrast-enhanced chest CT image. The dataset is divided into $870$ training subjects (with $22256$ annotated LNs), $100$ validation subjects (with $2600$ LNs), and $100$ testing subjects (with $2349$ LNs). All LNs were initially annotated using a $3$D bbox by two radiologists. For the cases where both radiologists marked the same LN, the annotations were accepted; while discrepancies were resolved by a third senior radiologist. The size of annotated LNs ranges from $5$mm to $40$mm. The original image size is $512$$\times$$512$$\times$$(312-672)$ with spacing $ 0.7$mm$\times$$0.7$mm$\times$$(1-1.5)$mm; After preprocessing, all the images were resampled to the spacing $1$mm$\times$$1$mm$\times$$1$mm.  

We implement our model with PyTorch and perform experiments on an NVIDIA A100 GPU. We use the Adam optimizer \cite{kingma2014adam}, with an initial learning rate of 0.001. Since we use the method of exponential decay of learning rate, the initial learning rate is then multiplied by $(1 - \frac{\text{epoch}}{{\text{max\_epoch}}})^{0.9}$. Due to the limitation of GPU memory, each volume is cropped into multiple patches with the size of 96$\times$160$\times$160 for training. 

\begin{table}
\caption{Performance of each method on Recall, Accuracy, F1 score. }
\label{table-recall}
\centering
\resizebox{0.56\textwidth}{!}{%
\begin{tabular}{cccc}
\hline
\multicolumn{4}{c}{FP = 2} \\ \hline
Method & Recall(\%) & Accuracy(\%) & F1(\%) \\ \hline
Swin-Det (w/o G) & 42.69 & 40.20 & 56.32 \\
CenterNet & 64.83 & 59.45 & 74.57 \\
Swin-Det (baseline) & 68.01 & 63.85 & 77.12 \\
nnU-Net (baseline) & 75.38 & 69.30 & 81.86 \\
SDF-Net (w/o ATF) & 70.40 & 65.01 & 78.53 \\
\textbf{SDF-Net (ours)} & \textbf{78.92} & \textbf{72.39} & \textbf{83.98} \\ \hline
\multicolumn{4}{c}{FP = 3} \\ \hline
Method & Recall(\%) & Accuracy(\%) & F1(\%) \\ \hline
Swin-Det (w/o G) & 55.45 & 50.54 & 61.36 \\
CenterNet & 69.90 & 61.54 & 76.18 \\
Swin-Det (baseline) & 72.45 & 64.02 & 77.99 \\
nnU-Net (baseline) & 79.20 & 70.00 & 82.35 \\
SDF-Net (w/o ATF) & 76.14 & 67.40 & 79.02 \\
\textbf{SDF-Net (ours)} & \textbf{83.27} & \textbf{73.35} & \textbf{84.62} \\ \hline
\multicolumn{4}{c}{FP = 4} \\ \hline
Method & Recall(\%) & Accuracy(\%) & F1(\%) \\ \hline
Swin-Det (w/o G) & 59.98 & 52.79 & 63.65 \\
CenterNet & 73.59 & 62.33 & 76.80 \\
Swin-Det (baseline) & 78.06 & 67.92 & 78.59 \\
nnU-Net (baseline) & 82.10 & 69.86 & 82.25 \\
SDF-Net (w/o ATF) & 80.38 & 68.67 & 81.23 \\
\textbf{SDF-Net (ours)} & \textbf{84.67} & \textbf{74.14} & \textbf{85.36} \\ \hline
\end{tabular}%
}
\end{table}

Given that LNs occupy a relatively small portion of the overall image volume, we adopt a balanced sampling strategy where 40\% probability that sampled image patches contain lymph
nodes, while the remaining patches are randomly sampled. The network is trained for 300 epochs. The loss weights are set to \(\mathrm{\sigma}_{1}\) = 2, \(\mathrm{\sigma}_{2}\) = 10, $\kappa = 1 $  \text{ and } $\mu = 0.1$.

\begin{figure}[h]
  \centering
  \includegraphics[width=0.5\textwidth]{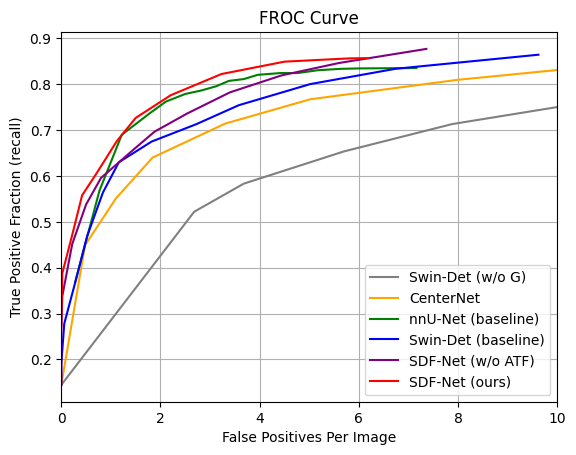}
  \caption{fROC Curve for all the comparative methods. The red one is our proposed method.}
  \label{froc}
\end{figure}

To evaluate the performance of our proposed method, we compared with the state-of-the-art deep learning algorithms, including \textbf{1)} \textbf{nnU-Net} \cite{isensee_nnu-net_2021}, \textbf{2)} \textbf{swin-Det} \cite{hatamizadeh_swin_2022-swin-unetr,zhou2019objects}, \textbf{3) CenterNet \cite{zhou2019objects}}, the anchor-free detection method. About the two baselines, nnU-Net (seg-path) is a cutting-edge segmentation approach while Swin-Det (det-path) is a novel detection algorithm that extends CenterNet \cite{zhou2019objects} by replacing the conventional CNN-based feature extraction with a transformer-based architecture from the Swin Transformer \cite{hatamizadeh_swin_2022-swin-unetr}.

For ablation studies, we evaluate our proposed transformer-based feature extraction, ATF and AVF modules of SDF-Net, and Gaussian kernel representation format. The methods include: 1) \textit{Swin-Det}, also as the det-path baseline, 2) \textit{SDF-Net (w/o ATF)}, SDF-Net without the ATF strategy, 3) \textit{SDF-Net (ours)}, which has the AVF, and 4) \textit{Swin-Det (w/o G)}, det-path without adaptive Gaussian kernel design. Here, we use the Free Response Operating Characteristic (fROC) curve, recall, accuracy, and F1 score to evaluate the detection performance.

\subsection{Experimental Results}

Fig. \ref{froc} shows the fROC curves of all compared methods. The recall of our proposed method is consistently higher than other methods. Notably, the Gaussian kernel design can substantially improve the detection performance. The ATF module also demonstrates improvements compared with the segmentation-only (nnU-Net) and detection-only (swin-Det) methods. 

Table. \ref{table-recall} shows the detailed performance. At different FP levels, the proposed method improves the detection performance in terms of recall, accuracy and F1 score. This effectively demonstrates the contribution of the proposed SDF-Net. 

\begin{figure}[h]
  \centering
  \includegraphics[width=1\textwidth]{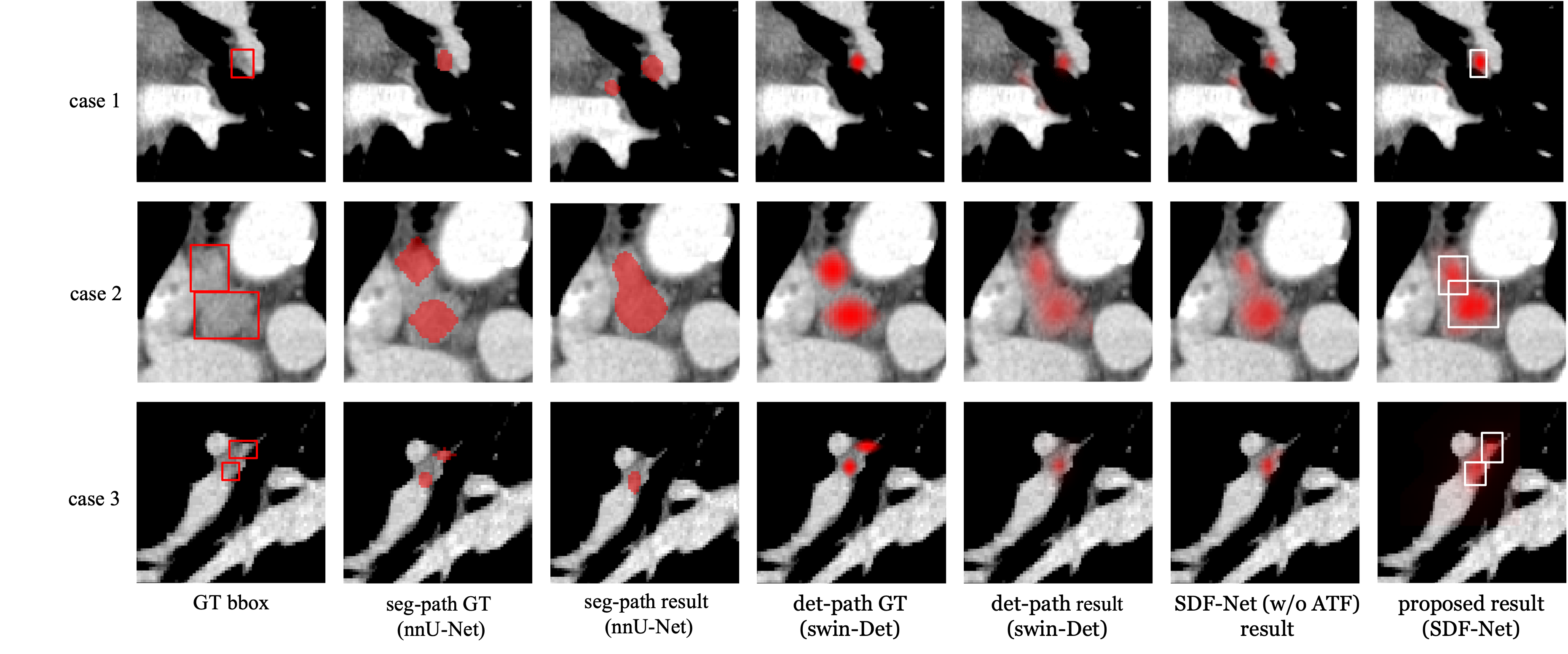}
  \caption{Visualization of results.}
  \label{noderesult}
\end{figure}

Fig. \ref{noderesult} visualizes LN detection results. The first column is ground truth bounding boxes. The second column illustrates the GT pseudo-mask drawn from bbox for segmentation-only model (nnU-Net), and the third column is the inference output of nnU-Net. The fourth column is the GT heatmap of detetcion-only model (swin-Det), i.e., the shape-adaptive Gaussian kernel, and the fifth column shows the heatmap inferred by swin-Det. The last two columns show the heatmap results by SDF-Net without ATF module, SDF(w/o ATF), and the whole proposed SDF-Net (proposed), respectively. The white boxes are the final detected bounding boxes.

Compared to the detection-only results (swin-Det), the proposed SDF-Net can effectively enhance small LN detection while reducing the FP, as shown in cases 1 and 3 in Fig. \ref{noderesult}. For the clustered LNs (case 2), the proposed method can accurately provide individual detection results, rather than the adhesion seen with segmentation-only model (nnU-Net). The proposed ATF module contributes to more accurate and distinct heatmap, leading to more precise detection results. 

\section{Conclusion}

We have proposed SDF-Net, a novel algorithm for detecting mediastinal lymph nodes in contrast-enhanced CT images. It integrates segmentation and detection features to address clustering issues and handle various LN shapes and sizes. A shape-adaptive Gaussian kernel further enhances detection. Experiments shown improvements over state-of-the-art methods.

\section*{Disclosure of Interests.}
The authors have no competing interests to declare that are relevant to the content of this article.

\bibliographystyle{unsrt}

\bibliography{ref} 

\begin{thebibliography}{10}

\bibitem{bouget2023mediastinal}
David Bouget, Andr{\'e} Pedersen, Johanna Vanel, Haakon~O Leira, and Thomas Lang{\o}.
\newblock Mediastinal lymph nodes segmentation using {3D} convolutional neural network ensembles and anatomical priors guiding.
\newblock {\em Computer Methods in Biomechanics and Biomedical Engineering: Imaging \& Visualization}, 11(1):44--58, 2023.

\bibitem{ren_faster-rcnn_2016}
Ross Girshick.
\newblock Fast {R-CNN}.
\newblock In {\em Proceedings of the IEEE International Conference on Computer Vision (ICCV)}, pages 1440--1448, 2015.

\bibitem{liu_ssd_2016}
Wei Liu, Dragomir Anguelov, Dumitru Erhan, Christian Szegedy, Scott Reed, Cheng-Yang Fu, and Alexander~C Berg.
\newblock {SSD}: Single shot multibox detector.
\newblock In {\em Proceedings of the European Conference on Computer Vision (ECCV)}, pages 21--37. Springer, 2016.

\bibitem{redmon_yolov3_2018}
Joseph Redmon and Ali Farhadi.
\newblock {YOLO}v3: An incremental improvement.
\newblock {\em arXiv preprint arXiv:1804.02767}, 2018.

\bibitem{law_cornernet_nodate}
Hei Law and Jia Deng.
\newblock {CornerNet}: detecting objects as paired keypoints.
\newblock In {\em Proceedings of the European Conference on Computer Vision (ECCV)}, pages 734--750, 2018.

\bibitem{zhou2019objects}
Xingyi Zhou, Dequan Wang, and Philipp Kr{\"a}henb{\"u}hl.
\newblock Objects as points.
\newblock {\em arXiv preprint arXiv:1904.07850}, 2019.

\bibitem{tian_fcos_2019}
Zhi Tian, Chunhua Shen, Hao Chen, and Tong He.
\newblock {FCOS}: A simple and strong anchor-free object detector.
\newblock {\em IEEE Transactions on Pattern Analysis and Machine Intelligence}, 44(4):1922--1933, 2020.

\bibitem{carion_end--end_2020-detr}
Nicolas Carion, Francisco Massa, Gabriel Synnaeve, Nicolas Usunier, Alexander Kirillov, and Sergey Zagoruyko.
\newblock End-to-end object detection with transformers.
\newblock In {\em European Conference on Computer Vision (ECCV)}, pages 213--229. Springer, 2020.

\bibitem{zhu_deformable_2021}
Xizhou Zhu, Weijie Su, Lewei Lu, Bin Li, Xiaogang Wang, and Jifeng Dai.
\newblock Deformable {DETR}: Deformable transformers for end-to-end object detection.
\newblock {\em arXiv preprint arXiv:2010.04159}, 2020.

\bibitem{isensee_nnu-net_2021}
Fabian Isensee, Paul~F Jaeger, Simon~AA Kohl, Jens Petersen, and Klaus~H Maier-Hein.
\newblock {nnU-Net}: A self-configuring method for deep learning-based biomedical image segmentation.
\newblock {\em Nature Methods}, 18(2):203--211, 2021.

\bibitem{hatamizadeh_swin_2022-swin-unetr}
Ali Hatamizadeh, Vishwesh Nath, Yucheng Tang, Dong Yang, Holger~R Roth, and Daguang Xu.
\newblock Swin {UNETR}: Swin transformers for semantic segmentation of brain tumors in {MRI} images.
\newblock In {\em International Conference On Medical Image Computing And Computer Assisted Intervention (MICCAI) Brainlesion Workshop}, pages 272--284. Springer, 2021.

\bibitem{ariji_segmentation_lymph-2022}
Yoshiko Ariji, Yoshitaka Kise, Motoki Fukuda, Chiaki Kuwada, and Eiichiro Ariji.
\newblock Segmentation of metastatic cervical lymph nodes from {CT} images of oral cancers using deep-learning technology.
\newblock {\em Dentomaxillofacial Radiology}, 51(4):20210515, 2022.

\bibitem{ronneberger_u-net_2015}
Olaf Ronneberger, Philipp Fischer, and Thomas Brox.
\newblock {U-Net}: Convolutional networks for biomedical image segmentation.
\newblock In {\em Proceedings of Medical Image Computing and Computer Assisted Intervention (MICCAI)}, pages 234--241. Springer, 2015.

\bibitem{li2019dice}
Xiaoya Li, Xiaofei Sun, Yuxian Meng, Junjun Liang, Fei Wu, and Jiwei Li.
\newblock Dice loss for data-imbalanced {NLP} tasks.
\newblock {\em arXiv preprint arXiv:1911.02855}, 2019.

\bibitem{wang2020multilay}
Feng Wang, Alberto Eljarrat, Johannes M{\"u}ller, Trond~R Henninen, Rolf Erni, and Christoph~T Koch.
\newblock Multi-resolution convolutional neural networks for inverse problems.
\newblock {\em Scientific Reports}, 10(1):5730, 2020.

\bibitem{lin_focal_nodate}
Tsung-Yi Lin, Priya Goyal, Ross Girshick, Kaiming He, and Piotr Doll{\'a}r.
\newblock Focal loss for dense object detection.
\newblock In {\em Proceedings of the IEEE International Conference on Computer Vision (ICCV)}, pages 2980--2988, 2017.

\bibitem{neubeck2006efficient-nms}
Alexander Neubeck and Luc Van~Gool.
\newblock Efficient non-maximum suppression.
\newblock In {\em 18th International Conference on Pattern Recognition (ICPR)}, volume~3, pages 850--855. IEEE, 2006.

\bibitem{kingma2014adam}
Diederik~P Kingma and Jimmy Ba.
\newblock Adam: A method for stochastic optimization.
\newblock {\em arXiv preprint arXiv:1412.6980}, 2014.

\end{thebibliography}

\end{document}